\documentclass[12pt]{article}
\usepackage{psfrag,epsf,color,subfigure}
\usepackage{enumerate}
\usepackage{natbib}
\usepackage[unicode=true,pdfusetitle,
   bookmarks=true, 
   bookmarksnumbered=false, 
   bookmarksopen=false,
   breaklinks=false, 
   backref=false, 
   colorlinks=false]{hyperref}
 \usepackage{mathtools}
\usepackage{amsthm,amsmath,amssymb,bbm,bm}
\usepackage{hyperref}
\usepackage{amssymb}
\usepackage{multirow}
\usepackage{array}
\usepackage{bm}
\usepackage{booktabs}
\usepackage{float}
\usepackage{multirow}
\usepackage[pdftex]{graphicx}
\usepackage{algorithm}
\usepackage{algorithmic}
\usepackage{placeins}
\usepackage{mathtools}
\usepackage{romannum}
\usepackage{mathrsfs}
\usepackage{dsfont}
\usepackage{relsize}
\usepackage{rotating}
\usepackage{enumitem}
\usepackage{setspace}
\usepackage{minitoc}
\usepackage{etoc}
\usepackage{graphicx}
\usepackage{subcaption}
\usepackage{listings}
\usepackage[font=small,labelfont=bf]{caption}
\setcitestyle{citesep={;}}
\usepackage[left=1in, right=1in, top=1in, bottom=1.2in]{geometry}
\usepackage{hyperref}

\usepackage{changes}
\usepackage[utf8]{inputenc} 
\usepackage[T1]{fontenc}    
\usepackage{hyperref}       
\usepackage{url}            
\usepackage{booktabs}       
\usepackage{amsfonts}       
\usepackage{nicefrac}       
\usepackage{microtype}      
\usepackage{xcolor}         

\usepackage{psfrag,epsf,color}
\usepackage{amssymb}
\usepackage{epsfig}
\usepackage{subfigure} 

\DeclareMathOperator*{\argmax}{arg\,max}
\DeclareMathOperator*{\argmin}{arg\,min}

\usepackage{authblk}
\usepackage{dsfont}

\newtheorem{thm}{Theorem}

\counterwithin{thm}{section}

\counterwithin{example}{section}
\counterwithin{rmk}{section}
\counterwithin{proposition}{section}

\counterwithin{corollary}{section}

\counterwithin{assumption}{section}
\newtheorem{definition}{Definition}[thm]
\counterwithin{definition}{section}
\newtheorem{assump}{Assumption}

\usepackage[page,header]{appendix}
\usepackage{titletoc}
\usepackage{lipsum}

\usepackage{caption} 
    \makeatletter
\def\@fnsymbol#1{\ensuremath{\ifcase#1\or *\or \mathparagraph\or
   \dagger\or \mathparagraph\or \|\or **\or \dagger\dagger
   \or \ddagger\ddagger \else\@ctrerr\fi}}
    \makeatother

\def\##1\#{\begin{align}#1\end{align}}
\def\$#1\${\begin{align*}#1\end{align*}}

\def\spacingset#1{\renewcommand{\baselinestretch}%
{#1}\small\normalsize} \spacingset{1}

\makeatletter
\newcommand{\vast}{\bBigg@{4}}
\newcommand{\Vast}{\bBigg@{5}}
\makeatother

\begin{document}

\title{\bf Stackelberg Batch Policy Learning}
\author{Wenzhuo Zhou}
\author{Annie Qu}
\affil{Department of Statistics, University of California Irvine}

\date{July 19, 2023}

\maketitle

\setcounter{page}{1}
\pagenumbering{arabic}

\begin{abstract}

Batch reinforcement learning (RL) defines the task of learning from a fixed batch of data lacking exhaustive exploration. Worst-case optimality algorithms, which calibrate a value-function model class from logged experience and perform some type of pessimistic evaluation under the learned model, have emerged as a promising paradigm for batch RL. However, contemporary works on this stream have commonly overlooked the hierarchical decision-making structure hidden in the optimization
landscape. In this paper, we adopt a game-theoretical viewpoint and model the policy learning diagram as a two-player general-sum game with a leader-follower structure. We propose a novel stochastic gradient-based learning algorithm: \textit{StackelbergLearner}, in which the leader player updates according to the total derivative of its objective instead of the usual individual gradient, and the follower player makes individual updates and ensures transition-consistent pessimistic reasoning. The derived learning dynamic naturally lends \textit{StackelbergLearner} to a game-theoretic interpretation and provides a  convergence guarantee to differentiable Stackelberg equilibria. From a theoretical standpoint, we provide instance-dependent regret bounds with general function approximation, which shows that our algorithm can learn a best-effort policy that is able to 
 compete against any comparator policy that is covered by batch data. Notably, our theoretical regret guarantees only require realizability without any data coverage and strong function approximation conditions, e.g., Bellman closedness, which is in contrast to prior works lacking such guarantees. Through comprehensive experiments, we find that our algorithm consistently performs as well or better as compared to state-of-the-art methods in batch RL benchmark and real-world datasets. 
\end{abstract}

\vspace{2mm}


\baselineskip=23pt

\section{Introduction}

Batch reinforcement learning (RL), also known as offline RL \cite{levine2020offline}, refers to the setting where policies are trained
using static, previously collected datasets. This presents an attractive paradigm for data reuse
and safe policy learning in many applications, such as healthcare \citep{murphy2001marginal,luckett2020estimating}, financial marketing \citep{theocharous2020reinforcement}, robotics \citep{thomas2015high} and education \citep{mandel2014offline}, as acquiring diverse or expert-quality data in these fields can be costly or practically unattainable. Recent studies have observed
that RL algorithms originally developed for the online or interactive paradigm perform poorly in the
offline case. This is primarily attributed to the distribution shift that arises over the course
of learning between the data-generating process of the batch dataset and the ones induced by the learned policy. Many prior works \citep{precup2000eligibility,ernst2005tree, antos2008learning,farahmand2016regularized}  crucially rely on a global data-coverage assumption and completeness-type function approximation condition in a technical sense. The former necessitates that the dataset to contain any state-action pair with a lower bounded probability so that the distributional shift can be well calibrated. The latter requires the function class to be closed under Bellman updates. Both assumptions are particularly strong and are likely to be violated in offline settings \citep{zanette2021provable}. Consequently, RL algorithms built depending on these assumptions may experience performance degradation and instability \citep{wang2021instabilities}. Thus, the development of novel algorithms
specialized for batch RL is of paramount importance in practical applications where only offline data is available. 

A major paradigm for algorithm design in batch RL is to incorporate pessimism 
for developing a worst-case optimality algorithm, which aims to maximize rewards in the worst possible Markov decision process (MDP) consistent with the offline dataset \citep{fujimoto2019off,xie2021bellman,zhan2022offline,fujimoto2021minimalist,kumar2020conservative,cheng2022adversarially}. In practice, these methods have generally been shown to be more robust when coverage assumptions are violated \citep{lee2021optidice}. Although prior worst-case optimality algorithms have relaxed the global coverage to a partial coverage condition, 
wherein the offline data is required to only cover some comparator policies \citep{uehara2022pessimistic}, all of these methods encounter challenges from both modeling and theoretical sides. From an algorithmic standpoint, most worst-case optimality-type algorithms often follow an actor-critic framework in order to perform pessimism \citep{zanette2021provable}. It naturally involves a hierarchical interaction between the actor and critic, i.e., a critic approximates the possible values of the
actor in unknown MDPs while concurrently learning an actor to optimize the pessimistic value based on the critic’s estimation. Unfortunately, existing works
have neglected this natural 
but the hidden hierarchical structure, and thus completely missed its potentially important utilization for convergence guarantee, solution concept, and pessimism reasoning in offline RL. From a theoretical viewpoint, most of the existing works require Bellman closedness for the function class \citep{xie2021bellman,cheng2022adversarially}, however slightly increasing the capacity of the function class could result in a new
class that does not have Bellman closedness anymore and suffers policy performance degradation. Even if the most recent works \citep{zhan2022offline} take a significant step towards relaxing Bellman closedness to realizability, the function class can capture the target ground-truth function, these algorithms are unable to provide a meaningful regret guarantee in the case of both global and partial coverage fails. Thus, we are interested in the following question: 

\textit{``Is it possible to take advantage of the implicitly hierarchical structure to formulate principled batch RL algorithms with strong theoretical
guarantees without any data coverage and Bellman-closedness conditions?''}

In this paper, we study this question from a game-theoretical perspective and provide an affirmative answer to the question. Our contributions can be summarized as follows. First, we explicitly model the 
hierarchical interaction as a two-player general-sum Stackelberg game:  the leader player optimizes its objective accounting for the anticipated response of the follower player, while the follower selects the best response to the leader's action to optimize its own objective. This further motivates us to design a new type of worst-case optimality algorithm --- \textit{StackelbergLearner}, in which the leader performs pessimistic evaluations based on the follower's actions on selecting a set of transition-consistent value functions with minimum weighed average Bellman error. Instead of independently optimizing leader-follower's objective using individual gradient dynamics, we solve the policy iteratively in a manner that reflects the hierarchical structure. This distinguishes our work from the prior works. In particular, we propose a Stackelberg gradient-based learning rule in which the leader player makes updates using
the total derivative of its objective defined using the implicit function theorem, and the follower player updates using the typical individual gradient dynamics. Consequently, \textit{StackelbergLearner} can efficiently learn a policy upon the game-theoretical solution concept, and perform a series of pessimism evaluations consistent with the Bellman equation without overly pessimistic reasoning.



Second, we study the stochastic gradient-based learning rule for sample-efficient learning of the Stackelberg equilibrium and provide a set of convergence analyses showing \textit{StackelbergLearner} is able to avoid saddle points in
the stochastic settings. To do this, we begin by developing a policy gradient theorem for the total derivative update, and characterize local Stackelberg equilibria (LSE) by locating critical points satisfying differential Stackelberg equilibria (DSE). Building upon this, we show the learned solution asymptotically converges to the LSE. To the best of our knowledge, this is the first provably convergent algorithm in the community of the worst-case optimality algorithm. 

Third, we show that, in offline settings with general function approximation, realizability is enough to learn a high-quality comparator policy. In other words, StackelbergLearner has a strong theoretical regret guarantee without requiring any data coverage and Bellman-closedness assumptions. This theoretical observation extends the applicability into a wide scope and is particularly meaningful in offline RL since the optimal policy is not usually covered by offline data. Moreover, we also perform a case study on a linear MDP with feature mapping. This yielded a more sample-efficient outcome, improving the previously best-known sample complexity factor from $\mathcal{O}(p)$ to $\mathcal{O}(p^{1/2})$ in a computationally tractable manner, where $p$ is for the dimension of linear MDP. For empirical studies, we conduct extensive experiments on synthetic, benchmark, and real datasets. The numerical results validate our theoretical results and demonstrate the superior performance of  StackelbergLearner.

The rest of the paper is organized as follows. Section \ref{sec:relate}
discusses the related works. Section \ref{sec:prelim} introduces the background for batch reinforcement learning and notations. Section \ref{sec:method} formally characterizes the Stackelberg policy learning dynamics.  A comprehensive theoretical study in convergence, regret of the proposed algorithm is provided in Section \ref{sec:theory}. Sections \ref{sec:exp} demonstrate the empirical performance of
our methods. We conclude
this paper with a discussion of possible future research directions. All technical proofs are provided in Appendix.

\section{Related Works}
\label{sec:relate}

Insufficient coverage of the dataset due to the lack of exhaustive exploration is known as the biggest challenge
in batch RL \citep{levine2020offline}. To tackle this issue, a number of works have been recently proposed
from both model-free \citep{liu2020provably,rashidinejad2021bridging,xie2021bellman,yin2021near,zhan2022offline, cheng2022adversarially} and model-based
perspectives \citep{yu2021combo,yin2021near,uehara2022pessimistic,matsushima2020deployment}. Broadly speaking, their methods hinge on the principle of pessimism and its variants, in the sense that the learned policy can avoid uncertain areas
not covered by batch data. For empirical works, deep batch RL algorithms \citep{fujimoto2019off, kumar2020conservative,kostrikov2021offline} exhibit impressive empirical performance,
but their theoretical consistency guarantees are limited to tabular MDPs. As a theoretical side, \cite{xie2021bellman,zanette2021provable,uehara2022pessimistic} propose worst-case optimality algorithms with PAC (probably approximately correct) guarantees under realizability, partial coverage,
and Bellman completeness. In terms of the learning framework, the minimax batch RL methods are related to ours. The minimax learning usually requires two function approximators, one for modeling the marginalized importance-weight function, and the other for modeling the value function. The desirable performance of this stream of method has led to a surge of interest within the RL community \citep{xie2019towards,uehara2020minimax,jiang2020minimax,nachum2019dualdice,liu2020understanding,shi2022minimax,zhou2022estimating}. The major differences between ours and the above-mentioned works are two-fold. First, we form our algorithm via a game-theoretical perspective which the prior works completely ignore, and derive the gradident-based learning rule via an implicit stochastic update over the total derivative of its objective
instead of the usual individual gradient. Second, we establish strong regret guarantees without relying on global coverage and even partial coverage conditions. We also relax the function approximation condition only requiring realizability rather than a strong Bellman-closedness condition.  

\section{Preliminaries}
\label{sec:prelim}

We consider an discounted infinite-horizon 
 Markov decision process (MDP) $\mathcal{M}=\{\mathcal{S}, \mathcal{A}, \mathds{P}, \gamma, r, s^{0}\}
$ \citep{sutton2018reinforcement}. Here,  $\mathcal{S}$ is the state space, $\mathcal{A}$ is the action space, $
\mathds{P}: \mathcal{S} \times \mathcal{A} \rightarrow \Delta(\mathcal{S})
$ is the Markov transition kernel on the  probabilistic simplex $\Delta$, $
r: \mathcal{S} \times \mathcal{A} \rightarrow [0, \bar{R}]
$ is the reward function for $\bar{R} \geq 0$, $\gamma \in [0,1)$ is the discounted factor, and $s^0$ is the initial state. A policy $\pi: \mathcal{S} \rightarrow  \Delta(\mathcal{A})$ induces a trajectory
\$
\{s^{0}, a^{0}, r^{0}, s^{1}, a^{1}, r^{1}, s^{2}, a^{2}, r^{2}, \ldots \},
\$
where $a^{t} \sim \pi(\cdot | s^{t}), r^{t}=r(s^{t}, a^{t})$ and $s^{t+1} \sim \mathds{P}(\cdot|s^{t}, a^{t})$ for any $t \geq 0$. The expected discounted return of the policy $\pi$ is defined as 
\$
J(\pi) = \mathbb{E}\Big[\sum_{t=0}^{\infty} \gamma^{t} r^{t} | \pi\Big].
\$
We call the return, when the trajectory starts with $(s, a)$ and all remaining actions are taken following $\pi$, as the true $q$-function $q^{\pi}: \mathcal{S} \times \mathcal{A} \rightarrow [0, V_{\max}]$. The true $q^{\pi}$ is the unique fixed point of the Bellman operator $\mathcal{B}^{\pi}
$, satisfying the Bellman equation \citep{puterman2014markov}: 
\$
\mathcal{B}^{\pi} q(s, a) \coloneqq r(s, a)+\gamma \mathbb{E}_{s^{\prime} \sim \mathds{P}(\cdot | s, a)}[q(s^{\prime}, \pi)],
\$
where $q(s^{\prime}, \pi)$ is denoted as shorthand for $
\mathbb{E}_{a^{\prime} \sim \pi\left(\cdot | s^{\prime}\right)}\left[q\left(s^{\prime}, a^{\prime}\right)\right]
$. Throught out this paper, we use the notation $\mathds{P}^{\pi}q(s,a) := \mathbb{E}_{s^{\prime} \sim \mathds{P}(\cdot | s, a)}\left[q\left(s^{\prime}, \pi\right)\right]$. It is also easy to see that $J(\pi) = q^{\pi}(s^{0},\pi)$. Moreover, the notion of 
the normalized discounted visitation of $\pi$ is defined as 
\$
d_{\pi}(s,a) \coloneqq (1-\gamma)\sum_{t=0}^{\infty} \gamma^{t} d_{\pi,t}(s,a),
\$
where $d_{\pi,t}$ is the marginal state-action distribution at the time-step $t$. In the batch RL setting, there exists an unknown offline data-generating distribution $\mu$ induced by behavior policies.  Despite the unknowns of $\mu$, we can observe a set of transition pairs, as offline dataset $\mathcal{D}_{1:n} \coloneqq \{s_i, a_i, r_i, s^{\prime}_i\}^{n}_{i=1}$ sampling from $\mu$. For a given policy $\pi$, the density-ratio (importance-weight), $\tau_{d_{\pi} / \mu}(s, a)
= d_{\pi}(s, a)/\mu(s, a)$, measures how effectively $\mu$ covers the visitation induced by $\pi$. The primary objective of offline policy optimization is to learn an optimal policy that maximizes the return, $J(\pi)$, using the offline dataset. 

Under the function approximation setting, we assume access to the policy and value function classes $\mathcal{Q}: \mathcal{S} \times \mathcal{A} \rightarrow \mathbb{R}$ and $\Pi: \mathcal{S} \times \mathcal{A} \rightarrow \Delta(\mathcal{A})$ which are utilized to capture $q^\pi$ and $\pi$, respectively. We also can construct a helper function class $\Omega: \mathcal{S} \times \mathcal{A} \rightarrow \mathbb{R}$ for modeling $\tau_{d_{\pi} / \mu}$ if necessary. The combination of
$\mathcal{Q}$, $\Pi$ (and $\Omega$) are commonly used in the literature of actor-critic, or minimax learning approach. We refer the readers to \citep{bertsekas1995neuro, konda2003onactor, haarnoja2018softac, uehara2020minimax} for details. For the most part of the paper, we do not make any structural assumptions on the function classes, making our
approach and guarantees applicable to general function approximation settings. We
now recall some standard assumptions on the expressivity of
the value function class $\mathcal{Q}$ which are needed for policy learning, particularly in an offline setting. 

\begin{assump}[Realizability]
For any policy $\pi \in \Pi$, we have 
$q^{\pi} \in \mathcal{Q}$. When this assumption holds approximately, we measure violation by
$
\inf _{q \in \mathcal{Q}} \sup _{ \rho}\mathbb{E}_{\rho}[\left(q(s,a)-\mathcal{B}^{\pi} q(s,a)\right)^{2}] \leq \varepsilon_{\mathcal{Q}} 
$,
where $ \varepsilon_{\mathcal{Q}} \geq 0$ and $\rho$ is some data distribution such that $\rho \in \{ d_{\widetilde{\pi}}: \widetilde{\pi} \in \Pi\}$.
\label{reliable_assum}
\end{assump}

Intuitively, Assumption \ref{reliable_assum} requires that for every $\pi \in \Pi$, there exists $q \in \mathcal{Q}$ that well-approximates $q^{\pi}$. Assumption \ref{reliable_assum} is a weaker form of stating $q^{\pi} \in \mathcal{Q}$, for any $\pi \in \Pi$. This realizability assumption is the same as the one made
by \cite{xie2021bellman} and is weaker than assuming a small
error in $L_{\infty}$ norm \citep{antos2008learning}. To our
knowledge, this is the minimum condition among existing works on batch RL with insufficient data coverage that has provided guarantees
under these standard assumptions for general function approximation. Notably, we do not require the class $\mathcal{Q}$ to be approximately closed under $\mathcal{B}^{\pi}$, the so-called Bellman-closedness assumption which is widely used
in existing RL theory but can be easily violated in real-world scenarios. 

When it comes to data exploration or coverage,  we say an offline dataset is well-explored if a well-designed behavior policy has been executed, allowing for a comprehensive exploration of the MDP environment. As a result, the dataset is likely to contain possibly all state-action pairs. This implicitly requires $\mu$ has the global coverage \citep{fujimoto2019off,uehara2022pessimistic}. In this context, the global coverage means that the density ratio-based concentrability coefficient, 
$\sup_{s,a} \{d_{\pi}(s, a) / \mu(s, a)\}$, is upper-bounded by a constant  $c \in \mathbb{R}^{+}$ for all policies $\pi \in \Pi$, where $\Pi$ is some policy class. This condition is frequently employed in batch RL \citep{antos2008learning,chen2019information,duan2020minimax}. However, in practice, this assumption may not hold true, as deploying an exploratory policy is a challenging task for large-scale RL problems. To solve this issue, we are naturally concerned with learning a good policy with strong theoretical guarantees that can compete against any arbitrarily covered comparator policy under much weaker conditions than global coverage.

\section{Stackelberg Policy Learning}\label{sec:method}

In this section, we introduce the notion of transition-consistent pessimism in the notation of the weighted average Bellman error and a new Stackelberg learning formulation of batch RL. We also highlight 
the hierarchical structure that exists in objective formulations and the Stackelberg learning rule under this design. This game-theoretical formulation deviates our algorithm from the prior works. 

A Stackelberg game \citep{von2010market} involves two players: one designated as the leader and the other as the follower. Each player aims to optimize their respective objectives, which depend on both their actions and those of their counterpart. In this game dynamic, the leader acts with the anticipation that the follower will respond optimally to the leader's move. Let $f_1\left(x_1, x_2\right)$ and $f_2\left(x_1, x_2\right)$ be the objective functions that the leader and follower want to minimize, respectively, where $x_1 \in X_1 \subseteq \mathbb{R}^{d_1}$ and $x_2 \in X_2 \subseteq \mathbb{R}^{d_2}$ are their decision variables or strategies and $x=\left(x_1, x_2\right) \in X_1 \times X_2$ is their joint strategy. The leader and follower aim to solve the following problems:
\$
& \text{Leader Player}: \quad \max_{x_1 \in X_1}\Big\{f_1\left(x_1, x_2\right) \mid x_2 \in \arg \min _{y \in X_2} f_2\left(x_1, y\right)\Big\}   \\
& \text{Follower Player}: \; \min _{x_2 \in X_2} f_2\left(x_1, x_2\right) .
\$
This contrasts with a simultaneous play game in which
each player is faced with an independent optimization problem. The policy learning algorithms we develop ensure that players adopt myopic update rules. These rules guide them to move in the direction of the steepest descent for their respective optimization problems.

Motivated by the design of worst-case optimality algorithms and pessimistic evaluation procedure in \citep{kumar2020conservative}, we propose to use the offline dataset to first compute a lower
bound on the value of each policy $\pi \in \Pi$, and then return the policy with the highest pessimistic
value estimate. While this architecture is the core part of many recent approaches \citep{fujimoto2019off,zanette2021provable, xie2021bellman}, our main novelties are the analysis of transition-consistent pessimism and the hierarchal policy learning design. In particular, drawing insights from the Stackelberg game, we explicitly cast the above-mentioned pessimistic evaluation procedure as the leader-follower problem: 
\#
& \text{Leader Player}: \quad \max_{\pi \in \Pi}\Big\{J(\pi,\underline{q^{\pi}})  \mid \underline{q^{\pi}} \in \argmin_{q\in \mathcal{Q}}\mathcal{L}(\pi,q)\Big\}  \notag \\
& \text{Follower Player}: \; \min _{q \in \mathcal{Q}} \mathcal{L}\left(\pi, q\right) := q(s^{0},\pi) + \lambda  \sup_{\tau \in \Omega}\underbrace{\mathbb{E}^2_{\mu}[\tau(s,a)(r(s,a) +\gamma q(s^{\prime},\pi)-q(s,a))]}_{\mathcal{L}^{0}(\pi,q)},
\label{lead-follow_optim}
\#
where the leader has the objective 
$J(\pi,q) := q(s^{0},\pi)$ and the follower has the objective $\mathcal{L}(\pi,q)$. Intuitively, the leader attempts to maximize the value estimate of $\underline{q^{\pi}}$ over some policy class $\Pi$, while the follower seeks a set of transition-consistent $q$-functions so that the leader only performs pessimistic evaluations on this set of $q$-functions. In the definition of the follower's objective, the $\sup_{\tau \in \Omega}\mathcal{L}^{0}(\pi,q)$ ensures the learned $q$-function is transition-consistent on offline data in terms of the notion of the sufficiently small weighted average Bellman error, and the $\mathcal{L}(\pi,q)$ promotes pessimism with $\lambda$ being the hyperparameter that softly controls their relative contributions in a regularization manner. In general, for a larger $\lambda$, the follower is dedicated to the feasible $q$-functions with transition consistency. In contrast, the follower allows for more tolerance for $q$-functions deviated from $q^{\pi}$ on data. Also, we remark that the use of \textit{soft} pessimism plays an important role in establishing the stochastic gradient-based learning rule which will be discussed later. 

Notably, the development and use of average weighted Bellman error $\mathcal{L}^{0}(\pi,q)$ is fundamental in relaxing  Bellman-closedness assumption. It distinguished our algorithm from the famous API/AVI-type algorithms, e.g., \citep{ernst2005tree,farahmand2016regularized,chen2019information,jin2021pessimism,xie2021bellman}, which rely on a squared or minimax Bellman error, and requires finding $f \in \mathcal{F}$ so that $\|f-\mathcal{B}f \|^{2}_{2,\mu} \approx 0$ to obtain $f \approx q^{\pi}$. Unfortunately, even with the infinite amount of data, the empirical estimate of $\|f-\mathcal{B}f \|^{2}_{L_{2}(\mu)}$, i.e., squared empirical Bellman error) is biased due to the appearance of unwanted conditional variance (the double-sampling issue). The API/AVI-type algorithms need a separate helper function class $\mathcal{G}$ for modeling $\mathcal{B}f$, and \cite{chen2019information} has shown that when the class $\mathcal{G}$ realizes the Bayes optimal regressor $\mathcal{B}f$ (Bellman-completeness condition), the estimation is consistent and unbiased. In comparison, $\mathcal{L}^{0}(\pi,q)$ is allowed to be estimated via a single but unbiased sample estimate, i.e., no double sampling issue, thanks to not using the squared loss. Therefore, there is no need to require the strong Bellman-closedness approximation assumption. 

Before formalizing learning rules to solve the formulated leader-follower problem, let us first discuss the
equilibrium concept studied in
in hierarchical play games. Specifically, we focus our
attention on local notions of the equilibrium concepts as is
standard in learning in games since the objective functions
we consider need not be convex or concave in general function approximation settings, e.g., using deep neural networks as approximators. 

\begin{definition}[Local Stackelberg Equilibrium \citep{bacsar1998dynamic}]\label{lse_def}
Consider $U_i \subset$ $X_i$ for each $i \in\{1,2\}$. The strategy $x_1^* \in U_1$ is a local Stackelberg solution for the leader if, $\forall x_1 \in U_1$,
$$
\sup _{x_2 \in R_{U_2}\left(x_1^*\right)} f_1\left(x_1^*, x_2\right) \leq \sup _{x_2 \in R_{U_2}\left(x_1\right)} f_1\left(x_1, x_2\right),
$$
where $R_{U_2}\left(x_1\right)=\left\{y \in U_2 \mid f_2\left(x_1, y\right) \leq f_2\left(x_1, x_2\right), \forall x_2 \in\right.$ $\left.U_2\right\}$. Moreover, $\left(x_1^*, x_2^*\right)$ for any $x_2^* \in R_{U_2}\left(x_1^*\right)$ is a local Stackelberg equilibrium on $U_1 \times U_2$. 
\end{definition}

While the existence of equilibria is beyond the scope of this paper, given its existence, we can characterize it based on sufficient conditions related to leader-follower's objective functions. We denote $D_i f_i$ as the derivative of $f_i$ with respect to $x_i, D_{i j} f_i$ as the partial derivative of $D_i f_i$ with respect to $x_j$, and $D(\cdot)$ as the total derivative, and let $D^{2}f_i$ be
the second-order total derivative. It observes that
$Df_1$ is the total derivative of $f_1$ with respect to $x_1$ given
$x_2$ is implicitly a function of $x_1$. This observation is aligned with the game strategy that the
leader plays under the expectation that the follower will
play the best response to $x_1$. In the following, we present and characterize the sufficient conditions
for a Local Stackelberg Equilibrium (LSE) as given in Definition \ref{lse_def}. 

\begin{definition}[Differential Stackelberg Equilibrium]\label{dse_def}
    The joint strategy $x^*=\left(x_1^*, x_2^*\right) \in X$ is a differential Stackelberg equilibrium if the following conditions are satisfied. 
    \$
    & (i) \; D f_1\left(x^*\right)=0, \; \text{and} \; D^2 f_1\left(x^*\right)> 0 \\
    & (ii) \; D_2 f_2\left(x^*\right)=0, \; \text{and} \; D_2^2 f_2\left(x^*\right)>0.
    \$
\end{definition}

We observe that the follower's reaction curve might not always be unique. Nonetheless, the sufficient conditions on a local Stackelberg solution $x$-i.e., $D_2 f_2(x)=0$ and $\operatorname{det}\left(D_2^2 f_2(x)\right) \neq 0$ --- ensure that $D f_1$ is well-defined, as indicated from the implicit function theorem \citep{abraham2012manifolds}. With sufficient conditions on LSE, we are able to study the stochastic gradient-based learning rule, which we name the Stackelberg learning rule, via the aforementioned game Jacobians. Specifically, we let 
$$
\omega_{\mathcal{S}}(x)=\left(D f_1(x), D_2 f_2(x)\right)
$$
be the vector of individual gradients. Then the vector field $\omega_{\mathcal{S}}(x)$, along with the derived Jacobian 
\$
J_{\mathcal{S}}(x)=\left[\begin{array}{cc}D_1\left(D f_1(x)\right) & D_2\left(D f_1(x)\right) \\ D_{21} f_2(x) & D_2^2 f_2(x)\end{array}\right],
\$
build the foundation of our stochastic gradient-based learning diagram, and characterizes sufficient conditions for a DSE. In particular, letting $\omega_{\mathcal{S}, i}$ be the $i$-th component of $\omega_{\mathcal{S}}$, the leader total derivative is $\omega_{\mathcal{S}, 1}(x)=D_1 f_1(x)-D_{21} f_2(x)^{\top}\left(D_2^2 f_2(x)\right)^{-1} D_2 f_1(x)$, and the follower individual derivative is $\omega_{\mathcal{S}, 2}(x) = D_2 f_2\left(x\right)$. Based on the first-order gradient-based sufficient conditions for DSE in Definition \ref{dse_def}, the stochastic Stackelberg learning rule we study for each player is given by
\#
x_{i, k+1}=x_{i, k}\pm \gamma_{i, k}\big(\omega_{\mathcal{S}, i}(x_k)+w_{k+1, i}\big),
\label{gradient_rule}
\#
where $\{w_{i, k}\}$ is player $i$ 's noise process and $\gamma_i,k > 0 $ is the step-size. While we can formulate the leader-follower optimization landscape in \eqref{lead-follow_optim} by following the above update rule for treating $f_{1}(x) = J(\pi, q)$ and $f_{2}(x) = \mathcal{L}(\pi, q)$, the estimation of total derivative $D J(\pi, q) $ is not straightforward due to a two-fold reason. First, the total derivative  $D J(\pi, q) $ involves multiple components that are required to be analyzed individually. Second, the nested inner maximization $\mathcal{L}^{0}(\tau,q)$ makes a minimax optimization scheme for the follower. To tackle these issues, we leverage the kernel embedding \citep{gretton2012kernel} to decouple the follower's minimax objective to a single-stage minimization objective which could be naturally connected to the stochastic gradient-learning rule \eqref{gradient_rule}. In particular, we model $\Omega$ in a bounded reproducing kernel Hilbert space (RKHS) equipped with a positive definite kernel $K(\cdot, \cdot)$, i.e.,  $\Omega_{\text{RKHS}}(C_K) \coloneqq \{\tau \in \text{RKHS}: \| \tau \|_{\text{RKHS}} \leq C_K \}$, where $\|\cdot\|_{\text{RKHS}}$ denotes the kernel norm and the constant $C_K >0$. This kernel representation allows the maximization problem $\sup_{\tau \in \Omega_{\text{RKHS}}(C_K)}\mathcal{L}^{0}(\tau,q)$ to have a simple closed-form solution $\tau^{*}$. Without loss of generality, we set $C_{k}=1$ for simplicity of notations. 
\#
\mathbb{E}_{\mu}^{1/2}\left[\Delta^q\left(q ; s, a, s^{\prime}\right) \Delta^q\left(q ; \tilde{s}, \tilde{a}, \tilde{s}^{\prime}\right)  K((s, a),(\tilde{s}, \tilde{a}))\right],
\label{kernel_loss}
\#
where $\Delta^q\left(q; s, a, s^{\prime}\right)=\left(r(s,a)+\gamma q\left(s^{\prime}, \pi\right)-q(s, a)\right)$, and $(\tilde{s},\tilde{a}, \tilde{s}^{\prime})$ is an independent copy of the transition pair $(s,a,s^{\prime})$. We also note that the square root operator is well-defined by the positive definite kernel $K(\cdot, \cdot)$. Therein, the follower's objective in \eqref{lead-follow_optim} is decoupled to a single minimization objective: 
\$
\min _{q \in \mathcal{Q}} q(s^{0},\pi) + \lambda \mathbb{E}_{\mu}\left[\Delta^q\left(q ; s, a, s^{\prime}\right) K((s, a),(\tilde{s}, \tilde{a}))\Delta^q\left(q ; \tilde{s}, \tilde{a}, \tilde{s}^{\prime}\right)\right].
\$
It then remains to tackle the issue of estimating the total derivative 
\#
D J(\pi, q) = D_{\pi} J(\pi, q)  - D_{q,\pi}\mathcal{L}(\pi, q)^{\top}\left(D^{2}_{q} \mathcal{L}(\pi,q)\right)^{-1} D_{q}J(\pi, q). 
\label{total}
\#
The individual gradient $D_{\pi} J(\pi, q)$ can be computed by the policy gradient theorem \citep{degris2012off}, i.e., 
$$
D_{\pi} J(\pi, q)=\mathbb{E}_{a \sim \pi(\cdot \mid s^{0})}\left[D_{\pi} \log \pi(a|s^{0}) q(s^{0}, a)\right].
$$
Furthermore, following a direct gradient calculation, we can obtain $
D_{q} J(\pi, q)=\mathbb{E}_{a \sim \pi(\cdot \mid s^{0})}\left[D_{q}q(s^{0}, a)\right]$. Similarly, we can also compute 
\$
D^2_{q}\mathcal{L}(\pi,q) = &  D^{2}_{q}q(s^{0},\pi) + 2\lambda \mathbb{E}_{\mu}\left[D \Delta^{q}(q;s, a, s^{\prime}) D \Delta^{q}(q;\tilde{s}, \tilde{a}, \tilde{s}^{\prime})^{\top} K((s, a),(\tilde{s}, \tilde{a})) \right] \\
& +  \Delta^{q}(q;s, a, s^{\prime}) K((s, a),(\tilde{s}, \tilde{a})) \big( \gamma D^{2}_{q}q(\tilde{s}^{\prime},\pi) - D^{2}_{q}q(\tilde{s}, \tilde{a}) \big).
\$
To compute $D_{q,\pi} \mathcal{L}(\pi, q)$ in \eqref{total}, we begin by obtaining $D_{\pi}\mathcal{L}(\pi, q)$ with the following follower-policy gradient theorem results
\$
D_{\pi}\mathcal{L}(\pi, q)  = &  \mathbb{E}_{a \sim \pi(\cdot \mid s^{0})}\left[D_{\pi} \log \pi(a|s^{0}) q(s^{0}, a)\right] \\
& + 2\lambda\mathbb{E}_{\mu}\Big[\mathbb{E}_{a \sim \pi(\cdot|s^{\prime})}[D_{\pi}\log\pi(a|s^{\prime})q(s^{\prime},a)]K((s, a),(\tilde{s}, \tilde{a}))\Delta^{q}(q;s, a, s^{\prime}) \Big].
\$
The above follower-policy gradient results allows us to easily compute $D_{q,\pi}\mathcal{L}(\pi,q) $ directly by the composite gradient $D_{q}(D_{\pi}\mathcal{L}(\pi, q))$. Given these derivations, the total derivative \eqref{total} can be estimated by
offline data, and our algorithm can be updated via the stochastic Stackelberg learning rule along with the individual derivative of the follower 
\$
D_{q} \mathcal{L}(\pi,q) = D_{q}q(s^{0},\pi) + 2\lambda \mathbb{E}_{\mu}\left[D \Delta^{q}(q;s, a, s^{\prime})\Delta^{q}(q;\tilde{s}, \tilde{a}, \tilde{s}^{\prime})^{\top} K((s, a),(\tilde{s}, \tilde{a}))\right]. 
\$
We summarize the update rules in Algorithm \ref{sto_rule}, where the true gradient is in replacement of the estimated stochastic gradient by offline data, denoted by $\widehat{D}$, which implicitly involves the noise term $w_{k+1, i}$ in \eqref{gradient_rule}. The leader-follower is assumed to have an unbiased estimator of the gradient appearing in their update rule. To estimate the stochastic gradients $\widehat{D}\circ$, we can leverage the U-statistic as the unbiased estimator \citep{shi2022minimax, zhou2022estimating}. 

\begin{algorithm}[H]
\setstretch{1.12}
	\caption{StackelbergLearner updating rule}
\label{sto_rule}
	\begin{algorithmic}[1]
	\STATE \textbf{Input} observed data $\mathcal{D}_{1:n}=\{(s_i,a_i,r_i,s_i^{\prime})\}^n_{i=1}$ and parameters $\gamma_{1,k},\gamma_{2,k}  $, and $\lambda$. 
				\STATE \textbf{For} $k=1$ to $\bar{K}$:
		\STATE \; Calculate $\widehat{D}J(\pi^{k},q^{k}) =  \widehat{D}_{\pi} J(\pi^{k}, q^{k})  - \widehat{D}_{q,\pi}\mathcal{L}(\pi^{k}, q^{k})^{\top}\big(\widehat{D}^{2}_{q} \mathcal{L}(\pi^{k},q^{k})\big)^{-1} \widehat{D}_{q}J(\pi^{k}, q^{k}) 
  $
	\STATE  \; Calculate the stochastic gradient $\widehat{D}_{q}\mathcal{L}(\pi^{k},q^{k})$  
 \STATE  \; Update $\pi^{k+1} \leftarrow \pi^{k} + \gamma_{1,k} \widehat{D}J(\pi^{k},q^{k})  $
  \STATE  \; Update $q^{k+1} \leftarrow q^{k} - \gamma_{2,k} \widehat{D}_{q}\mathcal{L}(\pi^{k},q^{k})  $
\end{algorithmic}
\end{algorithm}

For practical purposes, we extend the current learning rule with an implicit regularization. The total derivative in the Stackelberg gradient dynamics requires computing the inverse of follower Hessian $D_{q}^2\mathcal{L}(\pi,q)$. Under the function approximation setting, the value-function approximation usually relies on a non-linear function approximation, e.g., deep neural networks. Therefore, the optimization is highly non-convex, and  $\left(D_{q}^2\mathcal{L}(\pi,q)\right)^{-1}$ can be ill-conditioned. To account for this issue, instead of computing this term directly in the stochastic Stackelberg learning algorithms, we compute a regularized variant of the form $\left(D_{q}^2\mathcal{L}(\pi,q)+\beta I\right)^{-1}$. This regularization method can be interpreted as the leader viewing the follower as optimizing a regularized cost $\mathcal{L}(\pi,q)+(\beta/2)\left\|q\right\|^2$, while the follower actually optimizes $\mathcal{L}(\pi,q)$.

\section{Main Results}
\label{sec:theory}

\subsection{Game-Theoretical Convergence Analysis}
In this section, we provide convergence guarantees in stochastic gradient settings. First, we show that the policy learning algorithm, StackelbergLearner, is able to avoid saddle points. Second, we provide a local convergence
guarantee for the algorithm to a local Stackelberg equilibrium. Recall that, StackelbergLearne makes updates following the form
\$
\pi^{k+1}= & \pi^{k}+\gamma_{1, k}(DJ(\pi^{k},q^{k}) + w_{1, k+1}) \\
q^{k+1}= & q^{k}-\gamma_{1, k}(D_{q}\mathcal{L}(\pi^{k},q^{k}) + w_{2, k+1}) ,
\$
where $\gamma_{1, k}=o\left(\gamma_{2, k}\right)$ and $\left\{w_{i, k+1}\right\}$ is a noise process to indicate the stochastic gradient settings for each $i=1,2$.

Before we present our theoretical results, we first make the following standard assumptions. 

\begin{assump}\label{lip}
The maps $D J(\pi,q): \mathbb{R}^m \rightarrow \mathbb{R}^{m_1}$, $D_{q} \mathcal{L}(\pi,q): \mathbb{R}^m \rightarrow \mathbb{R}^{m_2}$ are Lipschitz, and $\left\|D J(\pi,q)\right\|<\infty$.    
\end{assump}

\begin{assump}\label{martin}
 For each $i \in\{1,2\}$, the learning rates satisfy $\sum_k \gamma_{i, k}=\infty$, $\sum_k \gamma_{i, k}^2<\infty$. The noise processes $\left\{w_{i, k}\right\}$ are zero mean, martingale difference sequences: given the filtration $\mathcal{F}_k=$ $\sigma\left(q^s, \pi^s, w_{1, s}, w_{2, s}, s \leq k\right),\left\{w_{i, k}\right\}_{i \in \mathcal{I}}$ are conditionally independent, $\mathbb{E}\left[w_{i, k+1} \mid \mathcal{F}_k\right]=0$almost surely, and $\mathbb{E}\left[\left\|w_{i, k+1}\right\| \mid \mathcal{F}_k\right] \leq$ $b_i\left(1+\left\|(\pi^{k},q^{k}))\right\|\right)$ almost surely for some constants $b_i \geq 0, i \in \mathcal{I}$.   
\end{assump}

Assumptions \ref{lip} and \ref{martin} are the standard assumptions for smoothness requirement and martingale property in stochastic approximation theory widely used in analyzing game-theory dynamics systems \citep{borkar2009stochastic, fiez2020implicit, zrnic2021leads}. With these assumptions, next we justify that StackelbergLearner is able to avoid saddle points in the following theorem. 

\begin{thm}[Saddle Points Escape]\label{non_conv}
Consider an MDP and the leader-follower has the objective $(J(\pi,q),\mathcal{L}(\pi,q))$ with the objective $J,\mathcal{L} \in C^l\left(\mathbb{R}^{m_1} \times \mathbb{R}^{m_2}, \mathbb{R}\right), l \geq 2$, where $C^l$ is the $l-$th order smooth function class. Suppose that for both leader, there exists a constant $c_1>0$ such that $\mathbb{E}\big[\left(w_{1, t} \cdot v\right)^{+} \mid \mathcal{F}_{1, t}\big] \geq c_1$ for every unit vector $v \in \mathbb{R}^{m_1}$, and it holds also for the follower that there exists a constant $c_2>0$ such that $\mathbb{E}\big[\left(w_{2, t} \cdot v\right)^{+} \mid \mathcal{F}_{2, t}\big] \geq c_2$ for every unit vector $v \in \mathbb{R}^{m_2}$. Under Assumptions \ref{lip} and \ref{martin}, our policy learning algorithm StackelbergLearner converges to strict saddle points of the game almost surely.    
\end{thm}

Theorem \ref{non_conv} is particularly meaningful in the learning process. First, while the saddle point guarantees a specific payoff for both players, it might not be the optimal result. StackelbergLearner has been shown to avoid the saddle points, which leads to a more desirable solution. Second, in the context of optimization, particularly in the setting with non-linear function approximation, saddle points can be problematic and get the learning algorithm stuck at saddle points. Theorem \ref{non_conv} indicates that StackelbergLearner is equipped with the ability to escape from saddle points, and thus become more flexible for function approximation conditions. 

In addition to non-asymptotical results, we also analyze the asymptotic convergence. These results, combined with the non-convergence guarantee in Theorem \ref{non_conv}, provide a comprehensive convergence analysis for the derived learning rule. The subsequent theorem provides a local convergence guarantee. It ensures that sample points will asymptotically converge to the locally asymptotic trajectories of the limiting singularly perturbed solution, and therefore attain stable DSE. For clarity regarding this theorem, remember that in a continuous-time dynamical system characterized by $\dot{z}=-g(z)$, a stationary point $z^*$
  of the system is deemed locally asymptotically stable if the spectrum of the Jacobian $-Dg(z)$ is in the open left half plane.


\begin{thm}[Asymptotic Convergence to Local Stackelberg Equilibria]\label{asym_conv}
Consider an MDP and the leader-follower has the objective $(J(\pi,q),\mathcal{L}(\pi,q))$ . Given a locally asymptotically stable differential Stackleberg equilibrium $\left(\pi^*, q^*\right)$ of the continuous-time limiting system $(\dot{\pi}, \dot{q})=\left(D J(\pi, q),-D_q L(\pi, q)\right)$, under Assumptions \ref{lip} and \ref{martin}, there exists a neighborhood $U$ for which the iterates $\left(\pi^k, q^k\right)$ of the discrete-time system in the update rule converge asymptotically almost surely to $\left(\pi^*, q^*\right)$ for $\left(\pi^0, q^0\right) \in U$. 
\end{thm}

This result effectively guarantees that the optimized solution locally converges to a stable,
game-theoretically meaningful equilibrium. That is, Theorem \ref{asym_conv} indicates that the solution is convergent to local Stackelberg equilibria and thus, unlike
simultaneous gradient descent without hierarchical learning, avoids converging to spurious
locally asymptotically stable points \citep{mazumdar2020gradient}. This convergence property not only stabilizes the learning procedure but provides a near-optimal guarantee in solving near-optimal policies.

\subsection{Regret Analysis}

In this section, we provide regret analyses of our policy learning algorithm, which reveals its advantages from a technical standpoint. Notably, to the best of our knowledge, Theorem \ref{main_thm_reg} is the first result of regret guarantee under only realizability \textit{without} requiring any data coverage or Bellman-closedness assumptions. Additionally, in contrast to most existing works that assume finite function classes, we carefully quantify the space complexities for infinite function classes (e.g., a class of real-valued functions generated by neural networks) using Pollard's pseudo-dimension \citep{pollard1990empirical} 
\begin{definition}[Pollard's pseudo dimension]
\label{pseudo_def}
    Let $\mathcal{F}$ be a class of functions from $\mathcal{X}$ to $\Re$. The pseudodimension of $\mathcal{F}$, written $D_{\mathcal{F}}$, is the largest integer $m$ for which there exists $\left(x_1, \ldots, x_m, \linebreak y_1, \ldots, y_m\right) \in \mathcal{X}^m \times \Re^m$ such that for any $\left(b_1, \ldots, b_m\right) \in\{0,1\}^m$ there exists $f \in \mathcal{F}$ such that
$$
\forall i: f\left(x_i\right)>y_i \Longleftrightarrow b_i=1.
$$
\end{definition}
 It notices that the pseudo-dimension is a generalization of the well-known VC dimension \citep{vapnik2015uniform}. In the following, we first introduce the routine assumptions before presenting the guarantees for our algorithm.

\begin{assump}[$L_{\infty}$-bounded $\mathcal{Q}$]
There exists a non-negative constant $V_{\max}  < \infty$, the function  $q(s,a) \in [0, V_{\max}], \, \forall q \in \mathcal{Q}, s \in \mathcal{S}$ and $a \in \mathcal{A}$. 
\label{q_bound}
\end{assump}

\begin{assump}[$L_{\infty}$-bounded  $\Omega_{\text{RKHS}}$]
There exists a non-negative constant $1 \leq \mathcal{C}^{\tau}_{\infty} < \infty$, the function $\tau(s,a) \in [0, \mathcal{C}^{\tau}_{\infty}], \, \forall \tau \in \Omega_{\text{RKHS}}, s \in \mathcal{S}$ and $a \in \mathcal{A}$.
\label{tau_bound}
\end{assump}

It is worth noting that we do not assume any form of Bellman-closedness and data-coverage assumption. For a learned policy $\widehat{\pi}$ of \eqref{lead-follow_optim}, we compete it with any policy $\pi \in \Pi$ and measure the $\gamma$-discounted infinite horizon value gap as
$$
\operatorname{Gap}(\pi, \widehat{\pi})=(1-\gamma)(J(\pi) -J(\widehat{\pi})).
$$

\begin{thm}\label{main_thm_reg}
Suppose Assumptions \ref{reliable_assum}-\ref{tau_bound} hold. Let 
\$
\mathcal{E}(\mathcal{Q},\Pi) = \big(1+e^{D_{\mathcal{Q}}+D_{\Pi}}\max\{D_{\mathcal{Q}},D_{\Pi}\}\big)^2(\mathcal{C}^{\tau}_{\infty})^{2(D_{\mathcal{Q}}+D_{\Pi})},
\$ 
for Pollard's pseudo-dimensions $D_{\mathcal{Q}}, D_{\Pi}$. And we set  
\$
\lambda = \widetilde{\mathcal{O}}\left(\sqrt[\leftroot{-1}\uproot{2}\scriptstyle 3]{\frac{n^2V_{\max}}{({\mathcal{C}}^{\tau}_{\infty}\ln \{\mathcal{E}(\mathcal{Q},\Pi) /\delta\})}}\right),
\$
then there exists a constant $\mathcal{C} \in [1,\mathcal{C}^{\tau}_{\infty})$, with probability at least $\geq 1-\delta$ , 
\$
 \operatorname{Gap}(\pi, \widehat{\pi}) \leq \; \widetilde{\mathcal{O}}\vast(
&  \min_{\rho \in \Theta(\rho,\mu,\mathcal{C}) } \Bigg\{ \mathbb{E}_{\left(d_{\pi}-\rho\right)^{+}}\bigg[\underbrace{\mathds{1}_{\mu=0}(\mathbb{I}-\gamma \mathds{P}^{\pi})\big(q^{\pi}_{\max}(s,a) - q^{\pi}_{\min}(s,a)\big)}_{\epsilon_{\text{off-supp}}} \\
&  \qquad \qquad \qquad \qquad \qquad \qquad +  \underbrace{\mathds{1}_{\mu>0}\mathfrak{M}({V_{\max},\gamma})\sqrt{\frac{\ln\{\mathcal{E}(\mathcal{Q},\Pi) /\delta\}}{n}} }_{\epsilon_{\text{bias}}}\bigg]\Bigg\} \\
&\underbrace{\mathcal{C}\sqrt[\leftroot{-1}\uproot{2}\scriptstyle 3]{\frac{\mathfrak{M}({V_{\max}})\ln\{\mathcal{E}(\mathcal{Q},\Pi)/\delta\}}{n}} }_{\epsilon_{\text{var}}} + \underbrace{\sqrt{\mathfrak{M}(\mathcal{C}^{\tau}_{\infty})\varepsilon_{\mathcal{Q}}}}_{\epsilon_{\text{model-mis}}} \vast),
\$
where $\Theta(\rho,\mu,\mathcal{C}) = \{\rho: \left\|\rho/\mu\right\|_{L_{2}(\mu)} \leq \mathcal{C} \}$, $q^{\pi}_{\max}:= \argmax_{q\in \mathcal{Q}_{\varepsilon_n}}q(s^{0},\pi)$ and $q^{\pi}_{\min}:= \argmin_{q\in \mathcal{Q}_{\varepsilon_n}}q(s^{0},\pi)$, and $\mathfrak{M}(\cdot)$ and $\widetilde{\mathcal{O}}$ denote the conditional constant terms, and big-Oh notation ignoring high-order terms, respectively.
\end{thm}

As stated in Theorem \ref{main_thm_reg}, we decompose the regret into the following error sources: the on-support uncertainty $\epsilon_{\text{var}}$, the on-support bias $\epsilon_{\text{bias}}$, the model-misspecification error on realizability $\epsilon_{\text{model-mis}}$, and the off-support error $\epsilon_{\text{off-supp}}$. Recall that we require $q^{\pi} \in \mathcal{Q}$ as in Assumption \ref{reliable_assum}, in fact, we can further relax the condition to requiring $q^{\pi}$ to be in the linear hull of $\mathcal{Q}$ \citep{uehara2020minimax}, which is more robust to the realizability error $\epsilon_{\text{model-mis}}$. 

In the following, we analyze the role of $\mathcal{C}$ in the regret bound from the two perspectives. First, we observe that the on-support statistical uncertainty $\epsilon_{\text{var}}$ is scaled by the constant $\mathcal{C}$, which measures the distribution shift between the implicit exploratory visitation $\rho$ and the offline data visitation $\mu$. In general, a small value of $\mathcal{C}$ controls 
the visitation $\rho$ to be closer to $\mu$, and thus reducing  $\epsilon_{\text{var}}$. Meanwhile, $\epsilon_{\text{bias}}$ depends on the probability mass of $(d_{\pi}-\rho)^{+}\mathds{1}_{\mu>0}$, representing the bias weighted by the probability mass difference between $d_{\pi}$ and $\rho$ within the on-support region of $\mu$. The small $\mathcal{C}$ potentially increase the mismatch between $d_{\pi}$ and $\rho$, and enlarges the on-support statistical bias $\epsilon_{\text{bias}}$. Consequently, there is a bias-variance tradeoff between on-support errors $\epsilon_{\text{bias}}$ and $\epsilon_{\text{var}}$, which is balanced via the constant $\mathcal{C}$. Second, the on-support error $\epsilon_{\text {var }}$ also demonstrates a behavior on making balance with respect to off-support error $\epsilon_{\text {off-supp }}$ via the distribution shift coefficient $\mathcal{C}$. Specifically, the set $\Theta(\rho,\mu,\mathcal{C})$ is potentially small if $\mathcal{C}$ is small. We are, therefore, limited in the choice of exploratory distributions $\rho$ that approximate visitation induced by $\widehat{\pi}$, and $\epsilon_{\text {off-supp}}$ could be large. In this case, $\epsilon_{\text {var}}$ could be small due to the distribution shift being well-controlled. On the contrary, for a large constant $\mathcal{C}$, the visitation measure induced by $\widehat{\pi}$ is well-approximate; however, it is more likely to suffer from a significant distribution shift from offline data distribution and yields large $\epsilon_{\text {var}}$.

When the offline dataset is under partial coverage, as indicated in \citep{uehara2022pessimistic}, it is necessary to provide a ``best-effort'' guarantee: learn a policy that is competitive to any good comparator policy if it is covered by offline data, not is just competitive optimal policy. This guarantee is particularly meaningful in practice because the optimal policy is typically not covered by batch data \citep{zhan2022offline}. Before we proceed to show StackelbergLearner is able to achieve this ``best-effort'' regret guarantee, we first formally define the partial coverage condition with respect to the density ratio concentrability as follows
\$
\left\|\frac{d_{\pi}(s,a)\mathds{1}_{\mu(s,a)=0}}{\mu(s,a)}\right\| < {\mathcal{C}}^{\tau}_{\infty}.
\$
We note that this single-policy partial coverage condition is much weaker than the global coverage condition which requires the above condition to hold for all polices. And this density ratio concentrability coefficient is standard and widely used in the literature \citep{chen2022offline,zhan2022offline}. Next, we state the near-optimal regret guarantee of our algorithm. 
\begin{thm}[Best-Effort Regret]\label{optimal_part_reg}
Suppose Assumptions \ref{reliable_assum}-\ref{tau_bound} hold. we set $\lambda$ as in Theorem \ref{main_thm_reg}, then for any good competitive policy $\pi^{*}$ such that
$
\left\|\frac{d_{\pi^{*}}(s,a)\mathds{1}_{\mu(s,a)=0}}{\mu(s,a)}\right\| < {\mathcal{C}}^{\tau}_{\infty}
$
with probability at least $\geq 1-\delta$, 
\$
 \operatorname{Gap}(\pi^{*}, \widehat{\pi}) \leq {\widetilde{\mathcal{O}}}\Bigg( \mathcal{C}\sqrt[\leftroot{-1}\uproot{2}\scriptstyle 3]{\frac{\mathfrak{M}({V_{\max}})\ln\{\mathcal{E}(\mathcal{Q},\Pi)/\delta\}}{n}} + \sqrt{\left( \mathcal{C}^{\tau}_{\infty} + 1\right) \varepsilon_{\mathcal{Q}}}\Bigg).
\$
\end{thm}

A close recent result to Theorem \ref{optimal_part_reg} is that of \citep{chen2022offline}, where they introduce a worst-case optimality algorithm grounded on a nontrivial performance gap condition. Their regret guarantee is only valid for the offline data covering the optimal policy. As we have discussed, it is typically that the optimal policy is not covered by the offline data. Regrettably, this result seems limited; the guarantee becomes void when the optimal policy is not covered by the offline data, a scenario that frequently arises. In contrast, Theorem \ref{optimal_part_reg} shows that StackelbergLearner can realize a near-optimal regret even in situations where the optimal policy is not covered in the offline data. In terms of the sample efficiency, Theorem \ref{optimal_part_reg} demonstrates that our algorithm enjoys a $\mathcal{O}(n^{-1/3})$ rate, which is much faster than $ \mathcal{O}(n^{-1/6})$ in the close work of \citep{zhan2022offline}, and competitive to $\mathcal{O}(n^{-1/3})$ of the recent work \citep{cheng2022adversarially}.

We now formalize another important advantage of our learning algorithm: baseline
policy improvement \citep{ghavamzadeh2016safe, laroche2019safe}, which can be viewed
as the special case when taking the competitive policy as the baseline policy. The robust baseline policy improvement is particularly important in the fields of medicine and finance with high-stake events \citep{tang2020clinician}. 

\begin{thm}[Robust Policy Improvement]
Suppose Assumptions \ref{reliable_assum}-\ref{tau_bound} hold with $\varepsilon_{\mathcal{Q}}=0$. we set $\lambda$ as in Theorem \ref{main_thm_reg}, and also we assume  $1 \in \Omega$ and the baseline policy $\pi_{b} = \in \Pi$ such that $d_{\pi_b} = \mu$, with probability at least $\geq 1-\delta$, $
 \operatorname{Gap}(\pi_{b}, \widehat{\pi}) \lesssim \sqrt[\leftroot{-1}\uproot{2}\scriptstyle 3]{\frac{\ln\{\mathcal{E}(\mathcal{Q},\Pi)/\delta\}}{n}}$.
\label{safe_imp} 
\end{thm}     

Theorem \ref{safe_imp} provides the robust policy improvement guarantee in the finite-sample regime, without relying on the Bellman-closedness condition. In contrast to the regular baseline policy improvement results in batch RL \citep{laroche2019safe,xie2021bellman}, the Bellman-closedness is necessary for their algorithm to achieve robust policy improvement. It can observe that their algorithm immediately
outputs degenerate solutions in the scenarios with complex transitions \citep{cheng2022adversarially}.

\subsection{A Cast Study with Feature Mapping}

In this section, we conduct a case study in linear MDPs with feature mapping (linear function approximation, \cite{zanette2021provable}). The motivation we studying the case of the linear MDP case is to position our theoretical result to the existing works as many prior works are derived based on the restrictive linear function approximation, e.g., \citep{jin2021pessimism,zanette2021provable}. The concept of the linear MDP is initially developed in the fully exploratory setting \citep{yang2020reinforcement}. Let $: \mathcal{S} \times \mathcal{A} \rightarrow \mathbb{R}^p$ be a $p$-dimensional feature mapping. We assume throughout that these feature mappings are normalized,
$\|\varphi(s,a)\|_{L_2} \leq 1$ uniformly for all  $(s,a) \in \mathcal{S} \times \mathcal{A}$. We focus on action-value functions that are linear in $\varphi$ and consider families of the following form:
\$
\mathcal{Q}_{\theta}:= \left\{(s, a) \mapsto\langle\varphi(s, a), \theta\rangle \mid\|\theta\|_{L_2} \leq c_{\theta}\right\},
\$
where $c_{\theta} \in [0, V_{\max}]$. For stochastic policies, we consider the soft-max policy class \citep{haarnoja2018softac}
\$
\Pi_{\omega}:= \{\pi_{\omega}(a|s)  \propto  e^{\langle\varphi(s, a), \omega\rangle} \mid\|\omega\|_{L_2} \leq c_\omega\},
\$
where $c_\omega \in (0, \infty)$. 

Before we present the regert guarantee, we make the assumption that the exact realizability holds for simplifying the analysis.

\begin{assump}[Exact Realizability]
\label{exact_reliable_assum}
For any policy $\pi \in \Pi_{\omega}$, we have 
$q^{\pi} \in \mathcal{Q}_{\theta}$. Then
\$
\inf _{q \in \mathcal{Q}} \sup _{ \rho}\mathbb{E}_{\rho}[\left(q(s,a)-\mathcal{B}^{\pi} q(s,a)\right)^{2}]=0.
\$
\end{assump}
In addition, inspired by \cite{agarwal2020optimality}, we can further relax the condition to characterize the partial coverage by combining the linear MDP structure. We formally define the concept of relative condition number (RCN) as follows:
\begin{definition}[RCN]
\label{def_relative_cond}
For any policy $\pi \in \Pi_{\omega}$ and behavior policy $\pi_{b}$ such that $d_{\pi_{b}}=\mu$,  the relative condition number is defined as
\$
\iota(d_{\pi},\mu) = \sup _{x \in \mathbb{R}^d} \frac{x^T \mathbb{E}_{ d_{\pi}}\left[\varphi(s, a) \varphi(s, a)^{\top}\right] x}{x^{\top} \mathbb{E}_{\mu}\left[\varphi(s, a) \varphi(s, a)^{\top}\right] x} .
\$
\end{definition}
With this definition, the partial coverage condition can be characterized through RCN, and we state it in the following assumption. 
\begin{assump}[Bounded RCN]
\label{rcn}
For any $\pi \in \Pi_{\omega}$,
$
\iota(d_{\pi},\mu) < \infty
$.
\end{assump}

Intuitively, this implies that as long as a high-quality comparator policy exists, which only visits the subspace defined by the feature mapping $\varphi$ and is covered by the offline data, our algorithm can effectively compete against it. This partial coverage assumption, in terms of RCN, is considerably weaker than density ratio-based assumptions \citep{uehara2022pessimistic}. In the following, we provide our main near-optimal guarantee in linear MDPs.

\begin{thm}
\label{linear_optim}
Suppose Assumption \ref{rcn} and \ref{tau_bound} hold. Set 
\$
& \lambda = \widetilde{\mathcal{O}}\bigg(\sqrt[\leftroot{-1}\uproot{2}\scriptstyle 3]{\frac{n^2}{(p\ln\{(e\sqrt{n}c_{\theta}c_\omega+1)/\delta\})^2}}\bigg), \\
& \Lambda(\kappa, \mathfrak{M}(\mathcal{C}^{\tau}_{\infty}), p, \pi, \mu) = \mathfrak{M}(\mathcal{C}^{\tau}_{\infty})\kappa p^2 \wedge \sqrt{\iota(d_{\pi},\mu)p}, 
\$ 
with probability at least $1-\delta$, 
 \$
 \operatorname{Gap}(\pi^{*}, \widehat{\pi}) \leq \widetilde{\mathcal{O}}\bigg(\Lambda(\kappa, \mathfrak{M}(\mathcal{C}^{\tau}_{\infty}), p, \pi, \mu)
\sqrt[3]{\frac{\mathfrak{M}({V_{\max}})p\ln \{(e\sqrt{n}c_{\theta}c_\omega+1)/\delta\}}{n}}\bigg),
 \$
where $\kappa = \text{trace}(\mathbb{E}_{\mu}[\varphi(s,a)\varphi(s,a)^{\top}])$.
\end{thm}

To the best of our knowledge, this is the first PAC-learnable result for an offline model-free RL algorithm in linear MDPs, requiring only realizability and single-policy concentrability. The regret bound we obtain is at least linear and, at best, sub-linear with respect to the feature dimension $p$. Our approach demonstrates a sample complexity improvement in terms of feature dimension compared to prior work by \citep{jin2021pessimism}, with a complexity of $\mathcal{O}(p^{1/2})$ versus $\mathcal{O}(p)$. It is worth noting that \citep{jin2021pessimism} only establishes results that compete with the optimal policy. Also, in the context of linear MDPs, they assume a global coverage from the offline data in contrast to the partial coverage condition we used. Meanwhile, while \cite{xie2021bellman} achieves a comparable rate as ours, the algorithm they proposed is not computationally tractable. In addition, they require a much stronger Bellman-completeness assumption and only work in some small action spaces.

\section{Experiments}
\label{sec:exp}

In this section, we evaluate the performance of our algorithm \textit{StackelbergLearner}, via comparing to the model-free batch RL baselines including CQL \citep{kumar2020conservative}, BEAR \citep{kumar2019stabilizing}, BCQ \citep{fujimoto2019off}, OptiDICE \citep{lee2021optidice}, and TD3+BC \citep{fujimoto2021minimalist}. We also compete with a popular model-based approach COMBO \citep{yu2021combo} as well while StackelbergLearner is a model-free algorithm for a comprehensive comparison. In addition to the policy performance evaluation, we also validate the theoretical rate of convergence on regret, and conduct ablation studies for the stability of StackelbergLearner. 

\subsection{Policy Performance}

In this section, we evaluate StackelbergLearner using policy performance as the criterion in comprehensive datasets including discrete and continuous control environments in synthetic datasets as well as well-studied offline RL benchmark datasets.

We first consider two synthetic environments, the CartPole benchmark environment from the OpenAI Gym \citep{brockman2016openai} and one simulated environment, to evaluate the policy performance of the proposed algorithm. For the first simulated environment setting, the system dynamics are given by
$$
\begin{aligned}
s^{t+1} & =\left(\begin{array}{cc}
0.75\left(2 a^{ t}-1\right) & 0 \\
0 & 0.75\left(1-2 a^{ t}\right)
\end{array}\right) s^{t}+\left(\begin{array}{cc}
0 & 1 \\
1 & 0
\end{array}\right)\odot
s^{t}{s^{t}}^{\top}\mathbb{I}_{2\times 1} + 
\varepsilon^t, \\
r^{t} & ={s^{t+1}}^{\top}\left(\begin{array}{l}
2 \\
1
\end{array}\right)-\frac{1}{4}\left(2 a^{ t}-1\right) + ({s^{t+1}}^{\top}s^{t+1})^{\frac{3}{2}}\odot\left(\begin{array}{c}
0.25 \\
0.5 
\end{array}\right),
\end{aligned}
$$
for $t \geq 0$, where $\odot$ denotes the Hadamard product, $\mathbb{I}$ is the identity matrix, the noise $\left\{\varepsilon^t\right\}_{t \geq 0} \stackrel{i i d}{\sim} N\left({0}_{2\times 1}, 0.25\mathbb{I}_{2\times 2}\right)$ and the initial state variable $s^{0} \sim N\left({0}_{2\times 1}, 0.25\mathbb{I}_{2\times 2}\right)$. The transition dynamic mainly follows the design in \cite{shi2020statistical}, but the reward function we consider here is more complex. In this setting, we consider a binary action space $a^{t} = \{0,1\}$. For the second environment, we take the CartPole environment from the OpenAI Gym \cite{brockman2016openai}, which is a widely-used benchmark RL dataset for policy evaluation. The state space in this environment is 4-dimensional and denoted as $s^{t} = (s^{t}_{[1]}, s^{t}_{[2]}, s^{t}_{[3]}, s^{t}_{[4]})$, comprising the position and velocity of the cart, as well as the angle and angular velocity of the pole. The action space consists of two actions, i.e., $\{0, 1\}$, which correspond to pushing the cart to the left or the right. We follow \cite{shi2022minimax} to apply a modified reward function to better differentiate values among various policies. The reward function is defined as:
\$
r^{t}=\left|2-\frac{s^{t}_{[1]}}{{s^{t}_{[1]}}({\text{clip}})}\right|\left|2-\frac{s^{t}_{[3]}}{{s^{t}_{[3]}}({\text{clip}})}\right|-1.
\$
Here, $s^{t}_{[1]}$ and $s^{t}_{[3]}$ represent the cart's position and the pole's angle, respectively. The terms ${s^{t}_{[1]}}({\text{clip}})$ and ${s^{t}_{[3]}}({\text{clip}})$ denote the thresholds at which the episode terminates (done = True) if either $|s^{t}_{[1]}| \geq {s^{t}_{[1]}}({\text{clip}})$ or $ |s^{t}_{[3]}| \geq {s^{t}_{[3]}}({\text{clip}})$ is satisfied. Under this definition, a higher reward is obtained when the cart is closer to the center and the pole's angle is closer to the perpendicular position.

In both settings, following \citep{uehara2020minimax}, we first learn a sub-optimal policy using DQN \citep{mnih2015human} and then apply softmax to its $q$-function, divided by a temperature parameter $\alpha$ to set the action probabilities to define a behavior policy $\pi_{b}$.
A smaller $\alpha$ implies $\pi_{b}$ is less explored, and thus the support of $\mu=d_{\pi_{b}}$ is relatively small. We vary different values of $\alpha$ for evaluating the algorithm performance in ``poor'', ``medium'' and ``well'' offline data coverage scenarios.  We use $\gamma=0.95$ and $\lambda=0.1$ with the sample size $n=1500$ in all experiments.
\begin{figure}
    \centering   \includegraphics[width=0.99\textwidth]{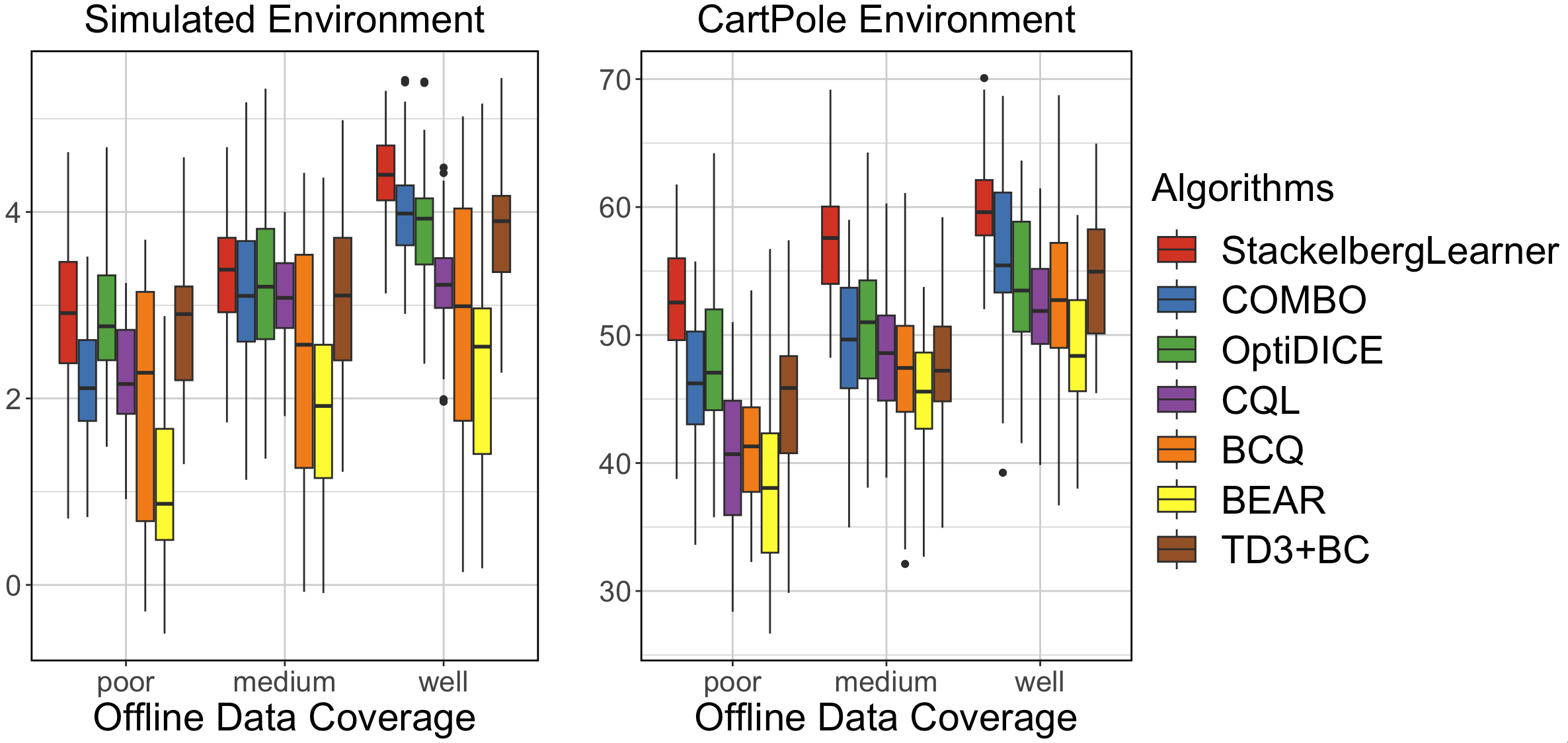}
    \caption{The boxplot of the discounted return over $50$ repeated experiments.}
     \label{fig:simu}
\end{figure}
Figure \ref{fig:simu} shows that StackelbergLearner consistently outperforms competing methods in different settings. The desirable performance mainly benefits from the advantages of the developed Stackelberg learning dynamics for convergence guarantee and the transition consistency achieved in the follower stage of StackelbergLearner. In addition, as illustrated by our theoretical results, StackelbergLearner does not rely on a strong function approximation condition: the Bellman-closedness. This helps StackelbergLearner outperform the competing methods, especially in a setting with poor offline data coverage where the Bellman-closedness is typically violated \citep{zhan2022offline}.


In addition to the empirical studies on policy performance using the synthetic data, we evaluate our algorithm in continuous MDPs using D4RL benchmark datasets \citep{fu2020d4rl}. Specifically, we make evaluations on Maze2D (3 tasks) and GymMuJoCo (12 tasks) domains in D4RL environments. We treat terminal states as absorbing states, adopting the absorbing state implementation in \citep{kostrikov2019imitation}. To design and deploy the behavior policy $\pi_{b}$ 
 , we use a tanh-squashed mixture of Gaussian policies to capture the multi-modality inherent in data sourced from heterogeneous policies as in \cite{lee2021optidice}. The normalized performance of our algorithm and competing algorithms for each task is presented in Table \ref{d4rl_res}. Each number in the table is the normalized score of the policy at
the last iteration of training, averaged over $5$ random seeds. We take the results of COMBO and OptiDICE from their original papers for GymMuJoCo, and run COMBO using author-provided implementations for Maze2D. The results of BCQ, BEAR methods from the D4RL original paper. In addition, CQL and TD3+BC are re-run to ensure a fair evaluation process for all tasks. Our algorithm achieves the best performance in 10 tasks and is comparable to the best baseline in the remaining tasks.
 Another noteworthy observation is that our algorithm overwhelmingly outperforms OptiDICE \citep{cheng2022adversarially} in almost all domains.  Although both StackelbergLearner and OptiDICE use the pessimism principle, OptiDICE uses a bouns-based pessimism without the Bellman consistency guarantee and thus causes over-pessimistic reasoning. In addition, StackelbergLearner takes advantage of the hierarchical structure of policy learning with a nice game-theoretical convergence property. This makes StackelbergLearner perform more stable with low standard deviation compared to all the competing baselines. 
 
\begin{table}[ht]
\footnotesize
\setlength{\tabcolsep}{0.05em}
\renewcommand{\arraystretch}{1.3}
\centering
\begin{tabular}{c|cccccccc}
\hline 
Tasks & StackelbergLearner & COMBO & BCQ & \ BEAR & OptiDICE & CQL & TD3+BC \\
\hline 
\hline
Maze2D &  &  & &  &  & &  & \\
\hline
maze2d-umaze & $96.5 \pm 7.8$ & $36.2 \pm 8.9$ & $12.8$ & $3.4$ & $\mathbf{111.0} \pm 8.3$ & $50.5 \pm 7.9$ &  $14.8 \pm 24.8$\\
        maze2d-medium   & $\mathbf{152.5} \pm 10.9$ & $47.8 \pm 12.6$ & $8.3$ & $29.0$ & $145.2 \pm 17.5$ & $28.6 \pm 9.2$  & $62.1 \pm 46.7$\\
        maze2d-large  & $\mathbf{187.8} \pm 15.2$ & $129.4 \pm 17.5$ & $6.2$ & $4.6$ & $155.7 \pm 33.4$  & $46.2 \pm 16.2$  & $88.6 \pm 16.4$\\   
\hline  
\hline  
GymMuJoCo &  &  & &  &  & &  & \\
\hline  
walker2d-random & $\mathbf{12.4} \pm 3.0$ & $7.0 \pm 3.6$ & $4.9$ & $7.3$ & $9.9 \pm 4.3$ & $4.7 \pm 4.5$ & $3.5 \pm 2.5$\\
walker2d-medium-expert & $101.2 \pm 5.4$ & $103.3 \pm 5.6$ & $57.5$ & $40.1$ & $74.8 \pm 9.2$ & $\mathbf{103.8} \pm 6.9$ & $101.4 \pm 8.8$ \\
walker2d-medium & $75.8 \pm 3.1$ & $72.9 \pm 4.8$ & $53.1$ & $59.1$ & $21.8 \pm 7.1$ & $73.2 \pm 4.2$ & $\mathbf{81.7} \pm 4.3$ \\
walker2d-medium-replay & $\mathbf{94.6} \pm 2.6$ & $56.0 \pm 8.6$ & $15.0$ & $19.2$ & $21.6 \pm 3.1$ & $20.8 \pm 2.8$ & $35.4 \pm 4.2$ \\
\hline  
hopper-random  & $\mathbf{18.7} \pm 1.5$ & $17.9 \pm 1.4$ & $10.6$ & $11.4$ & $11.2 \pm 1.1$ & $10.7 \pm 0.1$ & $14.1 \pm 2.2$ \\
hopper-medium-expert  & $\mathbf{117.8} \pm 0.5$ & $111.1 \pm 2.9$ & $110.9$ & $96.3$ & $111.5 \pm 0.6$ & $111.4 \pm 1.2$ & $111.4 \pm 1.3$\\    
hopper-medium  & $95.7 \pm 4.3$ & $97.2 \pm 5.2$ & $54.5$ & $52.1$ & $94.1 \pm 3.7$ & $74.3 \pm 5.8$ & $\mathbf{97.8} \pm  6.6$ \\
hopper-medium-replay  & $\mathbf{104.0} \pm 1.8$ & $89.5 \pm 2.2$ & $33.1$ & $33.7$ & $36.4 \pm 1.1$ & $32.6 \pm 4.9$ & $45.4 \pm 8.8$ \\ 
\hline  
halfcheetah-random  & $\mathbf{44.3} \pm 1.2$ & $38.8 \pm 3.7$ & $2.2$ & $25.1$ & $11.6 \pm 1.2$ & $26.7 \pm 1.4$ & $26.7 \pm 4.8$\\
halfcheetah-medium-expert  & $87.4 \pm 3.1$ & $90.0 \pm 5.6$ & $64.7$ & $53.4$ & $91.1 \pm 3.7$ & $66.7 \pm 8.9$ & $\mathbf{96.4} \pm 3.9$\\   
halfcheetah-medium  & $\mathbf{60.1} \pm 0.3$ & $54.2 \pm 1.5$ & $40.7$ & $41.7$ & $38.2 \pm 0.1$ & $37.2 \pm 0.3$ &  $25.8 \pm 2.4$\\
halfcheetah-medium-replay  & $\mathbf{58.9} \pm 1.1$ & $55.1 \pm 1.0$ & $38.2$ & $38.6$ & $39.8 \pm 0.3$ & $41.9 \pm 1.1$ & $46.3 \pm 1.7$\\ 
        \hline 
\end{tabular} 
  \caption{Normalized performance over $5$ random seeds of StackelbergLearner compared with
 the model-free and model-based baseline in the D4RL benchmark tasks \citep{fu2020d4rl}. For each task in \{hopper, walker2d, halfcheetah\} of MuJoCo continuous controls, the dataset is gathered in
the following ways: \textit{random}: the dataset is generated by a randomly initialized policy in each task. \textit{medium}: the dataset is generated by using the policy trained by \citep{haarnoja2018softac} with early stopping; \textit{medium-replay}: the \textit{replay} dataset consists of the samples gathered during training the policy; \textit{medium-expert}: the dataset is given by using the same amount of expert trajectories and suboptimal trajectories, where
those suboptimal ones are gathered by using either a randomly uniform policy or a medium-performance policy. For the tasks in the Maze2D environment, the complexity of the maze increases with the order of \textit{maze2d-umaze}, \textit{maze2d-medium} and \textit{maze2d-large}.}
\label{d4rl_res}
\end{table}

\subsection{Regret Rate of Convergence and Ablation Study}

In this section, we design a simulated environment that is able to help us accurately calculate the convergence rate of the regret. These results are consistent with our theoretical developments in regret bounds and convergence analyses. in Section \ref{sec:theory}. 
Next, we describe the data-generating process in our simulation study. We follow \cite{chen2023steel} to consider the mean reward function $r(s, a)=(a-\boldsymbol{W} s)^{\top} \boldsymbol{M}(a-\boldsymbol{W} s)$. Here, the state 
$s$ has a dimension of $p=6$, while the action $a$ has a dimension of $|\mathcal{A}|=5$. The parameter matrix $\boldsymbol{W}$ belongs to $\mathbb{R}^{|\mathcal{A}| \times p}$, and $\boldsymbol{M}$ is a $|\mathcal{A}| \times |\mathcal{A}|$ negative definite matrix. Each entry of $\boldsymbol{W}$ is uniformly sampled from the range $(0,1)$. The matrix $\boldsymbol{M}$ is formulated as $\boldsymbol{M}=-\boldsymbol{M}_0^{\top} \boldsymbol{M}_0$, where each entry of $\boldsymbol{M}_0$ follows a standard normal distribution. The initial reward is generated by $r^{1} = r\left(s^0, a^0\right)+\varepsilon_0$, with the independent white noise $\varepsilon_0$
  distributed as $\mathcal{N}(0,1)$. Consequently, the optimal policy can be analytically defined as $\pi^{*}(s)=s^{\top} \boldsymbol{W}$ for every state $s \in \mathcal{S}$, independent of the state distribution.

In terms of generating the training dataset, the state $s^{0}$ is uniformly sampled from $[0,1.5]^{p}$, mirroring the reference distribution. We keep the values of learned policies in this study initialed over the identical distribution to the training one. Actions are drawn following a behavior policy  $\pi^b$, such that $a^{0}=\boldsymbol{W} s^{0}+\varepsilon^{\mathcal{A}}$ where $\varepsilon^{\mathcal{A}} \sim \mathcal{N}\left(0, \sigma_0^2 \boldsymbol{I}_{|\mathcal{A}|}\right)$. Thus, the increase in $\sigma_0$
  suggests a behavior policy that deviates more from the optimal policy and also implies a more expansive exploration of the action space and better coverage for the training offline data.

\begin{figure}[H]
    \centering
\includegraphics[width=0.78\textwidth]{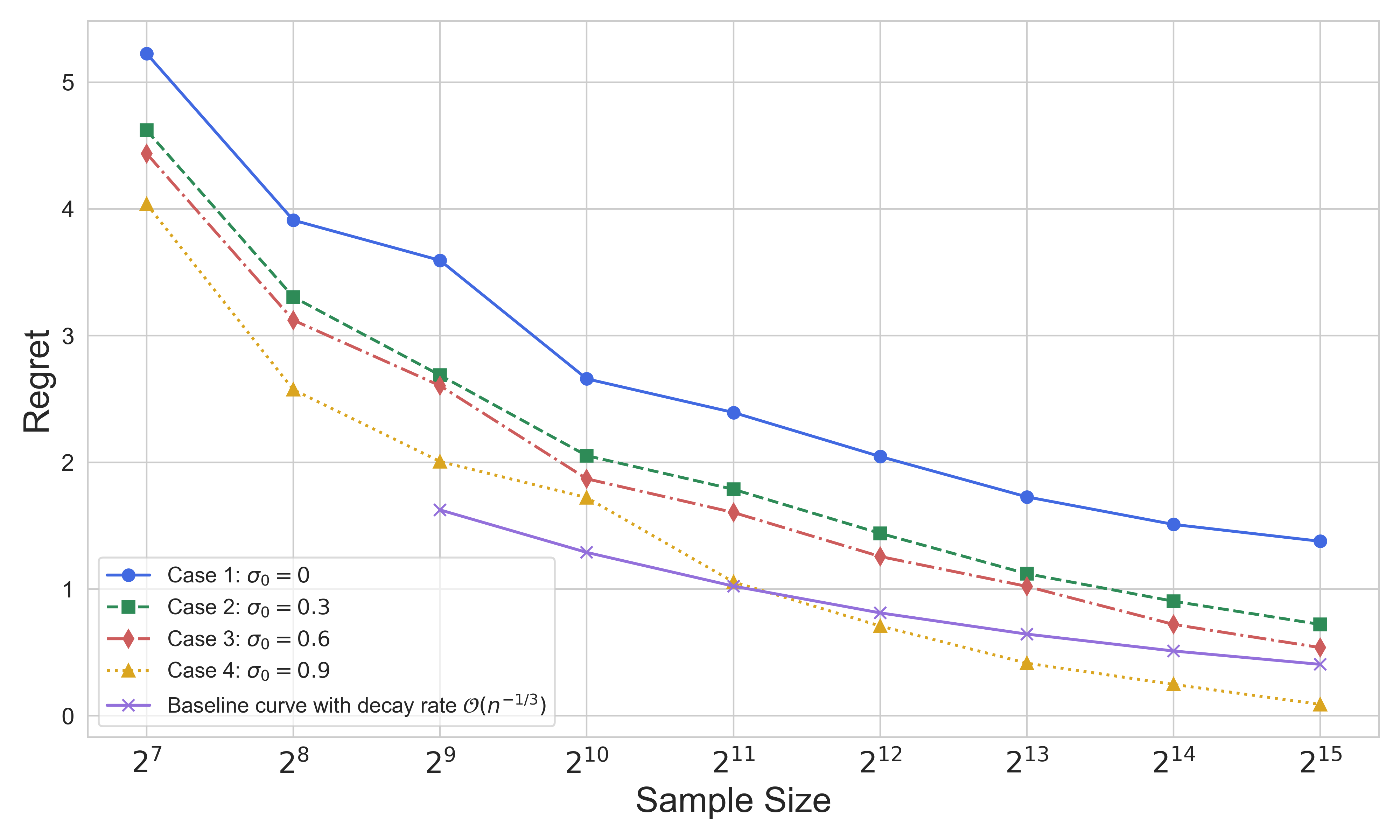}
    \caption{Convegence rate of the near-optimal regret (compete to the optimal policy) on the synthetic dataset with
different offline data coverage $\sigma_0$.}
    \label{fig:regret}
\end{figure}
Under this setting, we can accurately compute the regret due to an analytical optimal policy form. We study the regret rate of convergence for  StackelbergLearne, by increasing the training sample size $n$. Recall that $\sigma_0$ influences the offline data coverage. A thoroughly explored action space within the offline dataset is advantageous for policy learning. This breadth provides a more extensive set of transition pairs, facilitating a better evaluation of the values associated with a vast majority of candidate policies. Conversely, when $\sigma_0$
  is small (meaning the behavior policy is close to the optimal policy), the value estimation for the optimal policy might be precise. However, the empirical value evaluation of a sub-optimal candidate policy could be involved with significantly high uncertainties, potentially leading to unintended over-estimation consequences. In Figure \ref{fig:regret}, we observe that our method yields a desired convergence with the regret decaying to zero of the order $\mathcal{O}(n^{-1/3})$, which validates the results in Theorem \ref{optimal_part_reg} and \ref{linear_optim}. The rate of convergence is almost identical to $\mathcal{O}(n^{-1/3})$ across varied data coverage scenarios, which shows that StackelbergLearner is able to consistently realize a small regret in well, and poor offline data coverage. 

\begin{figure}[t]
    \centering
\includegraphics[width=0.78\textwidth]{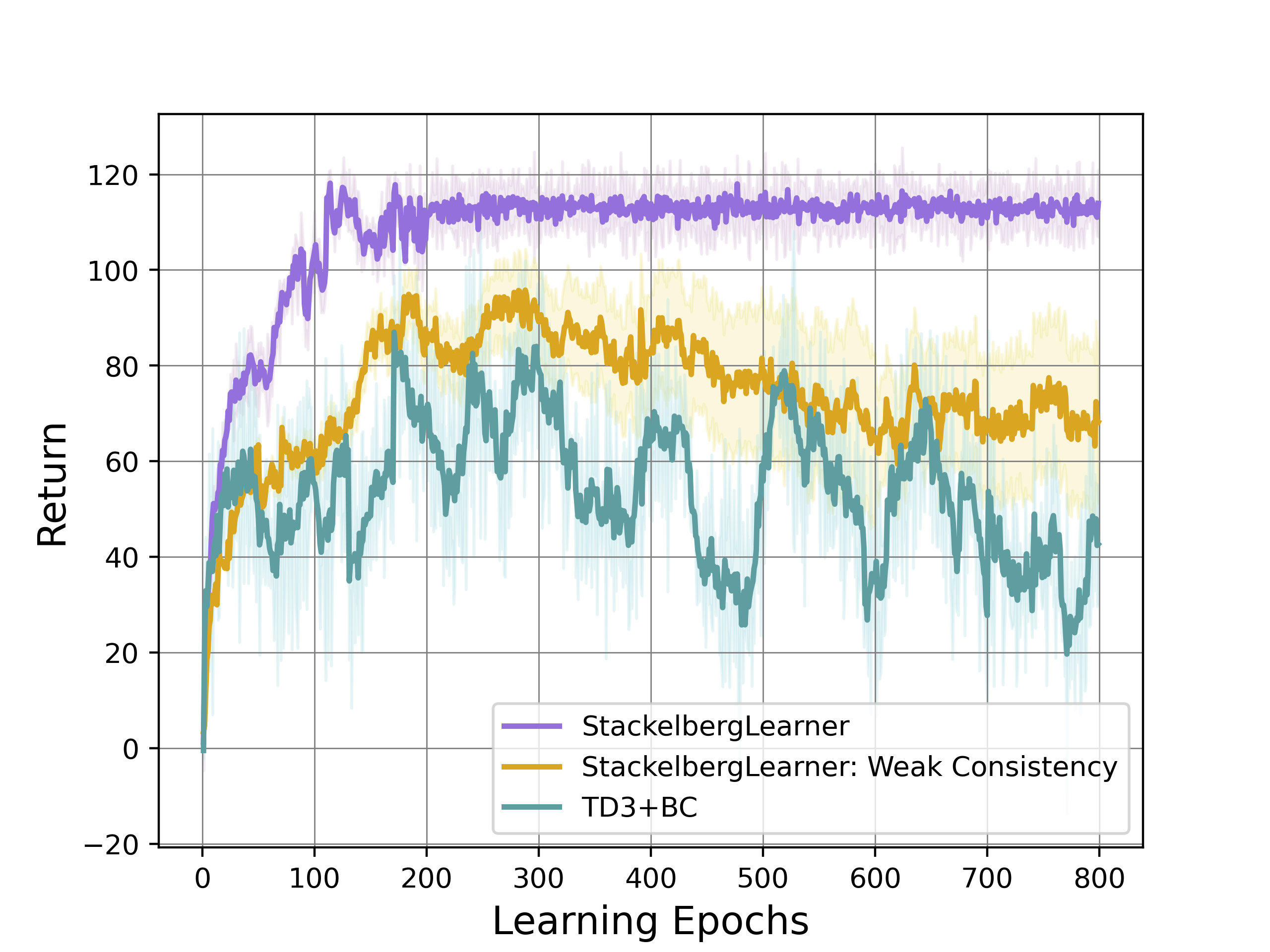}
    \caption{Abalation study on the learning stability and convergence over $5$ random seed experiments on the environment \textit{hopper-medium-replay}. The plot shows the policy performance across learning epochs for three algorithms: StackelbergLearner, \textit{StackelbergLearner: weak consistency}, and TD3+BC.
    The algorithm \textit{StackelbergLearner: weak consistency} is yielded by setting an extremely small $\lambda$, which intentionally downgrades the effect of the transition consistency property in the follower stage of StackelbergLearner.}
    \label{fig:ab}
\end{figure}

At last, we conduct ablation studies on the effect of the transition-consistency ensured in the follower stage of StackelbergLearner in terms of stability, convergence and policy performance. We also make a comparison to the recent baseline TD3+BC \citep{fujimoto2021minimalist}
. To this end, we study the policy performance across the learning epochs for three algorithms:   \textit{StackelbergLearner}, \textit{StackelbergLearner: weak consistency} by setting a very small $\lambda$ for downgrading the effect from follower, and TD3+BC in Figure  \ref{fig:ab}. The observations summarized from the plot are as follows. First, \textit{StackelbergLearner} achieves the best policy performance with a stable learning process indicated by the minor standard deviation of policy performance across epochs. The desirable performance is blessed by the transition-consistency property and the saddle points avoidance and asymptotic convergence to a local StackelbergLearner equilibrium as stated in Theorems \ref{non_conv} and \ref{asym_conv}. Second, for the algorithm \textit{StackelbergLearner: weak consistency} wherein the transition-consistency is almost removed, Figure \ref{fig:ab} shows the learning curve is still relatively stable but presents a performance degradation, especially in the later training stage. The performance degradation indicates that \textit{StackelbergLearner: weak consistency} suffers the over-estimation issue on the value function updating, which is probably due to the failure in guaranteeing transition-consistent value function set. Third, the baseline TD3+BC ignores the hierarchical learning structure and thus overlooks the nice convergence property if modeling from a game-theoretical viewpoint. Therefore, it observes that the training process of TD3+BC is quite unstable with highly fluctuated variability in offline settings.


\subsection{Real Data Analysis}

In this section, we apply and evaluate the StackelbergLearner on the Ohio Type 1 Diabetes (OhioT1DM) mobile health study, as detailed in \citep{marling2020ohiot1dm}. This study encompasses data from six patients with type 1 diabetes, capturing eight weeks of life-event data, including health status measurements and insulin injection dosage, per patient. Given the unique glucose dynamics exhibited by each patient based on the 
preliminary study, we follow \cite{zhu2020causal} to treat each patient's data as an independent dataset. Within this framework, the data from each day represents a distinct trajectory.

 We summarize the collected measurements over $60$-min intervals with the maximum horizon length as $24$. After dropping missing samples and outliers, each patient's dataset contains approximately $n=360$ transition pairs. The state variable $s^{t}$ is set to be a three-dimensional vector including the average blood glucose levels $s^{t}_{[1]}$, the average heart rate $s^{t}_{[2]}$ and the total carbohydrates $s^{t}_{[3]}$ intake during the period time $[t-1,t]$. We follow \citep{rodbard2009interpretation} to define the reward as the average of the index of glycemic control between time $t-1$ and $t$, which is used for indicating the health status of the patient's glucose level. That is
\$
{r^{t}}=-\frac{\mathbb{I}({s^{t}_{[1]}}>140)|{s^{t}_{[1]}}-140|^{1.10} + \mathbb{I}({s^{t}_{[1]}}<80)({s^{t}_{[1]}} - 80)^{2}}{30},
\$
which implies that reward $r^{t}$ is non-positive and a larger value is preferred \citep{zhou2022estimating}. In this real data analysis, we target to study a dose-finding problem, where the action space is a continuous dose level of the insulin injection. 

\begin{table}[ht]
\centering
\footnotesize
\renewcommand{\arraystretch}{1.5} 
\begin{tabular}{@{}lccccccc@{}} 
\toprule
\textbf{Patient ID} & \textbf{StackelbergLearner} & \textbf{COMBO} & \textbf{BCQ} & \textbf{BEAR} & \textbf{OptiDICE} & \textbf{CQL} \\
\midrule
$540$ & $\mathbf{20.7} \pm \mathbf{0.4}$ & $17.8 \pm 1.0$ & $14.6 \pm 0.5$ & $12.9 \pm 0.6$ & $18.2 \pm 0.8$ & $16.8 \pm 0.4$ \\
$544$ & $\mathbf{13.5} \pm 1.2$ & $10.1 \pm 1.4$ & $7.8 \pm 2.4$ & $6.2 \pm \mathbf{0.9}$ & $10.6 \pm 1.7$ & $9.0 \pm 0.9$ \\
$552$ & $8.2 \pm 0.8$ & $7.1 \pm 0.6$ & $6.0 \pm 0.4$ & $5.3 \pm 0.9$ & $\mathbf{8.4} \pm 0.8$ & $7.0 \pm \mathbf{0.3}$ \\
$567$ & $\mathbf{37.2} \pm \mathbf{1.2}$ & $30.9 \pm 2.1$ & $24.6 \pm 1.5$ & $26.0 \pm 1.3$ & $29.1 \pm 2.3$ & $26.8 \pm 1.3$ \\
$584$ & $\mathbf{33.4} \pm 1.7$ & $27.3 \pm 1.4$ & $20.6 \pm 1.3$ & $23.2 \pm 1.7$ & $28.0 \pm 1.8$ & $22.0 \pm \mathbf{1.3}$ \\
$596$ & $\mathbf{6.8} \pm \mathbf{0.7}$ & $4.4 \pm 0.7$ & $4.1 \pm 0.8$ & $3.0 \pm 1.2$ & $5.0 \pm 1.0$ & $4.9 \pm 0.9$ \\
\bottomrule
\end{tabular}
\caption{The Monto Carlo discounted returns for the baseline policy improvement over 50 repeated experiments.}
\label{real_data}
\end{table}

To evaluate the policy performance, since the underlying data-generating process in the environment is unknown, we follow \cite{luckett2020estimating} to use the Monte Carlo approximation of the return to evaluate the policy performance. Moreover, we randomly pick up $20$ trajectories from each individual based on available trajectories and repeat this procedure $50$ times to report the 
mean and standard deviation of the Monto Carlo returns in Table \ref{real_data}. Except for Patient $552$ , StackelbergLearner consistently outperforms the competing baselines for the majority of patients, showing its promising utility in real-world scenarios.
 


\section{Conclusion}
\label{sec:con}

We study batch policy learning with insufficient data coverage due to offline learning. We propose a StackelbergLearner with a leader-follower structure to cast an implicit hierarchical decision-making diagram, which can be efficiently solved with solutions to differentiable Stackelberg equilibria. We characterize the derived stochastic gradient-based learning rule 
and establish strong theoretical regret guarantees not relying on any
data-coverage and Bellman closedness conditions but only realizability. We further perform a case study in linear MDP, showing StackelbergLearner is a sample-efficient learner which improves the existing best regret results from a sub-linear rate to a linear rate depending on feature dimension $p$. Empirically, we conduct extensive experiments to evaluate the policy performance, convergence and stability, and validate theoretical results, demonstrating the superior performance of StackelbergLearner. Also, we remark that our algorithm has a great advantage over the existing methods \citep{cheng2022adversarially,xie2021bellman}: they propose to solve 
adversarial optimization problems and hence computational cost.

One of the potential future directions is to explore environments with unobservable confounders. How to yield an 
offline hyperparameter selection scheme across different
datasets or yield an automatically selected rule conditioned on the model error would be interesting. Moreover, how to extend StackelbergLearner in partial observable transition settings, i.e., POMDP, is also an interesting research direction. We leave these problems for future work.

\bibliographystyle{asa}
\bibliography{mycite}

\end{document}